\documentclass{article}

\usepackage{PRIMEarxiv}
\usepackage[utf8]{inputenc} 
\usepackage[T1]{fontenc}    
\usepackage{hyperref}       
\usepackage{url}            
\usepackage{booktabs}       
\usepackage{amsmath,amssymb,amsfonts}       
\usepackage{nicefrac}       
\usepackage{microtype}      
\usepackage{lipsum}
\usepackage{fancyhdr}       
\usepackage{graphicx}       
\graphicspath{{media/}}     
\usepackage{multirow}
\usepackage{subcaption}
\usepackage{threeparttable}

\title{Predicting Lung Cancer Patient Prognosis with Large Language Models
}

\author{
  Danqing Hu \\
  Zhejiang Lab \\
  Hangzhou, Zhejiang, China\\
  \texttt{hudq@zhejianglab.com} \\
   \And
  Bing Liu\\
  Peking University Cancer Hospital\\
  and Institute \\
  Beijing, China\\
  \texttt{liubing983811735@126.com} \\
   \And
  Xiang Li\\
  Peking University Cancer Hospital\\
  and Institute \\
  Beijing, China\\
  \texttt{xiangli@bjmu.edu.cn} \\
   \And
  Xiaofeng Zhu \\
  Zhejiang Lab \\
  Hangzhou, Zhejiang, China\\
  \texttt{andy.zhu@zhejianglab.com} \\
  \And
  Nan Wu \\
  Peking University Cancer Hospital\\
  and Institute \\
  Beijing, China\\
  \texttt{nanwu@bjmu.edu.cn} \\
}

\begin{document}
\maketitle

\begin{abstract}

Prognosis prediction is crucial for determining optimal treatment plans for lung cancer patients. Traditionally, such predictions relied on models developed from retrospective patient data. Recently, large language models (LLMs) have gained attention for their ability to process and generate text based on extensive learned knowledge. In this study, we evaluate the potential of GPT-4o mini and GPT-3.5 in predicting the prognosis of lung cancer patients. We collected two prognosis datasets, i.e., survival and post-operative complication datasets, and designed multiple tasks to assess the models' performance comprehensively. Logistic regression models were also developed as baselines for comparison. The experimental results demonstrate that LLMs can achieve competitive, and in some tasks superior, performance in lung cancer prognosis prediction compared to data-driven logistic regression models despite not using additional patient data. These findings suggest that LLMs can be effective tools for prognosis prediction in lung cancer, particularly when patient data is limited or unavailable.

\end{abstract}


\section{Introduction}

Lung cancer continues to be the leading cause of cancer-related mortality worldwide \cite{Sung2021}. For patients diagnosed with early-stage lung cancer, surgical resection offers the only potential cure \cite{Howington2013}. Accurately predicting post-operative prognosis, including complications and survival, is essential for clinicians to optimize treatment strategies. Such predictions enable clinicians to implement early interventions and develop personalized follow-up plans, ultimately improving patient outcomes.

To predict prognosis accurately, researchers often rely on data-driven methods to develop prediction models. Early efforts involved combining patients' clinical features with statistical approaches to build predictive models and nomograms \cite{HuangLiu2016,LiCui2017,WangWu2016,WangWang2022}. For censored survival data, the Cox proportional hazards model was commonly used for survival prediction \cite{HuangLiu2016,LiCui2017}. To uncover nonlinear relationships among clinical variables, machine learning techniques such as random forests, random survival forests, support vector machines, and survival support vector machines were employed, which enhanced model performance \cite{KourouExarchos2015,AdeoyeTan2021,YuTan2020,XiaoMo2022,HuZhang2022,XueLi2021,AlsinglawiAlshari2022}. With the rapid advancement of deep learning, these techniques have become the leading and most effective tools for developing prognosis prediction models \cite{KatzmanShaham2018,HuLi2020,MukherjeeZhou2020,TsengWang2020,HoferLee2020,MeyerZverinski2018,HuLiuLi2023}. Despite their success in cancer prognosis, these data-driven methods have a critical limitation: they require substantial amounts of clinical data, limiting their applicability in scenarios where such data is limited or unavailable.

Recently, large language models (LLMs) like ChatGPT \cite{OpenAI2024} and GPT-4 \cite{Achiam2023} have gained significant global attention for their advanced text-generation capabilities. Trained on extensive datasets, these models can tackle new tasks using zero-shot, one-shot, or few-shot prompts without requiring updates to their parameters \cite{Brown2020}. The integration of reinforcement learning from human feedback (RLHF) further enhances LLMs, enabling them to generate content that is both safe and consistent with human expectations  \cite{Ouyang2022}. This advancement has led to a significant shift in natural language processing (NLP) research and is beginning to impact clinical prediction research \cite{Tang2023,HuChen2024,Doshi2024,HuZhang2024,HuLiu2023}.

By utilizing medical knowledge acquired from extensive training data, LLMs have shown promise in diagnosing and assessing patient prognoses. While some studies have explored the use of LLMs for predicting clinical prognoses such as readmission, length of hospital stay, and mortality, their performance has generally not surpassed that of traditional machine learning models \cite{Chung2024,Glicksberg2024,Changho2024,ZhuWang2024}. In this study, we aim to leverage the large language models to predict the prognosis of lung cancer patients. We collected two prognosis datasets, i.e., survival and post-complication data, from the Department of Thoracic Surgery II, Peking University Cancer Hospital and Institute. GPT-4o mini and GPT-3.5 were employed to make prognosis predictions based on patients' clinical data. Our experimental results demonstrate that LLMs with a zero-shot prompt can outperform baseline logistic regression models in lung cancer prognosis prediction.

\section{Materials and methods}

\subsection{Prognosis datasets}

\begin{itemize}

\item \textbf{Survival dataset (SD)} We gathered a survival dataset from 1,277 lung cancer patients who underwent pulmonary resections at Peking University Cancer Hospital. This dataset includes patient demographics, medical history, surgical information, pathological characteristics of the primary tumor and lymph nodes, and the pathological TNM stage. To ensure the accuracy and reliability of the data, clinicians manually recorded all clinical information. Follow-up data, including the event indicator (survival or not) and the timing of the event or censoring, were also documented by the clinicians.

\item \textbf{Post-operative complication dataset (PCD)} We also included 593 lung cancer patients who underwent curative pulmonary resection in a separate post-operative complication dataset. All patients were over 70 years old and had no distant metastases. This dataset captures patient demographics, disease history, pre-operative examinations, surgical information, and the pathological TNM stage. Post-operative complications recorded include atelectasis, pleural effusion, asthma attack, lung infection, pulmonary embolism, deep vein thrombosis, arrhythmia, and angina. Clinicians manually collected all clinical data and complications from electronic medical records.

\end{itemize}

Ethical approval for this study was granted by the Ethics Committee of Peking University Cancer Hospital (2022KT128) before this study.

\begin{figure}[h]
\centering
\includegraphics[width=\textwidth]{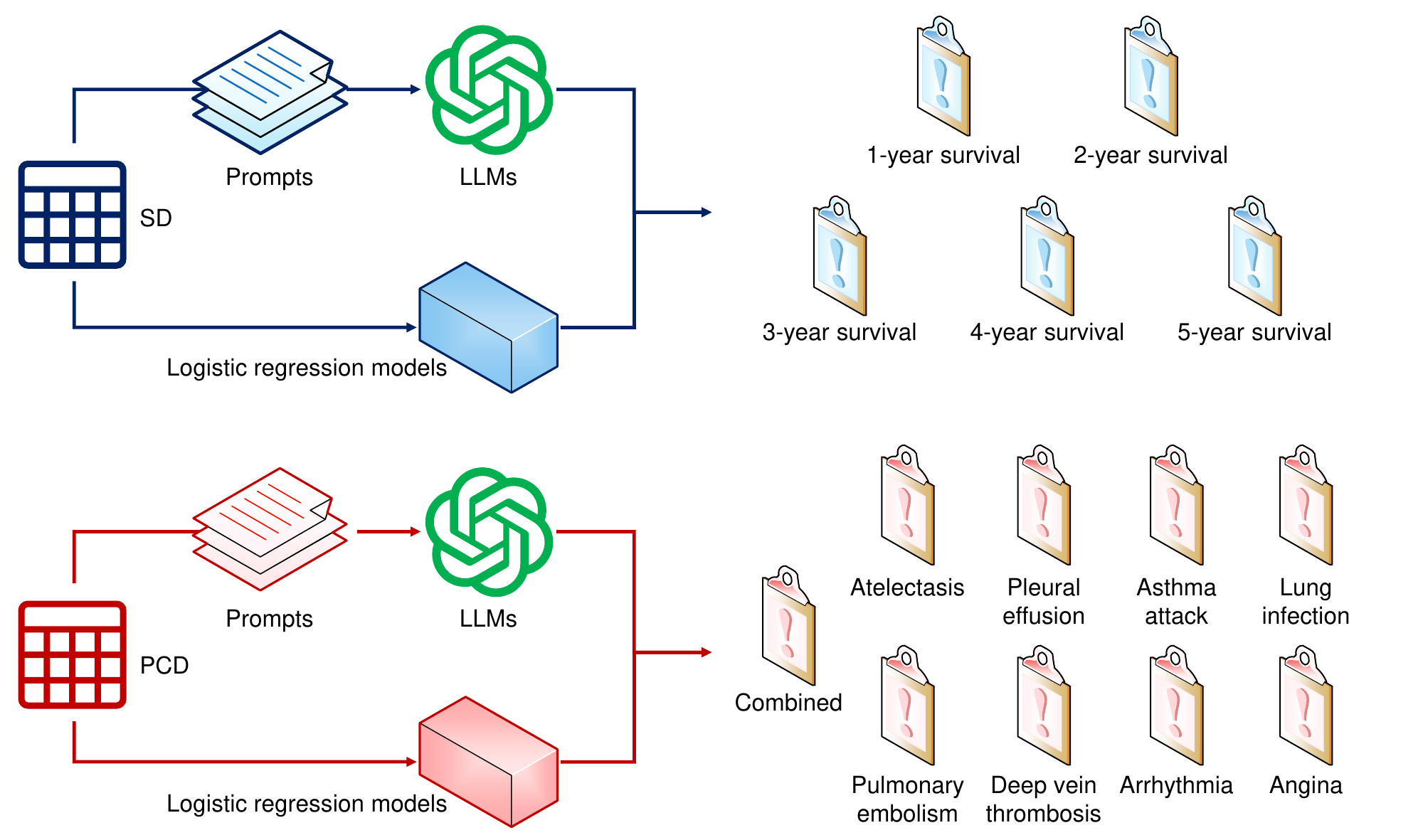}
\caption{Overall study design.}
\label{figure1}
\end{figure}

\subsection{Study design}

This study aims to leverage large language models (LLMs) to predict the prognosis of lung cancer patients. The overall study design is presented in Figure \ref{figure1}. We utilized the survival and post-operative complication datasets to generate prompts as input for the LLMs to obtain the predicted results. Additionally, we trained logistic regression models on these datasets to serve as the baselines. For survival prediction, we establish 5 prediction tasks, i.e., 1-year, 2-year, 3-year, 4-year, and 5-year survival predictions. For complication prediction, we predicted the risk of each of the eight complications and also combined these complications into a single label to predict.

\begin{figure}[h]
\begin{centering}
    \begin{subfigure}{0.45\textwidth}
        \includegraphics[width=\textwidth]{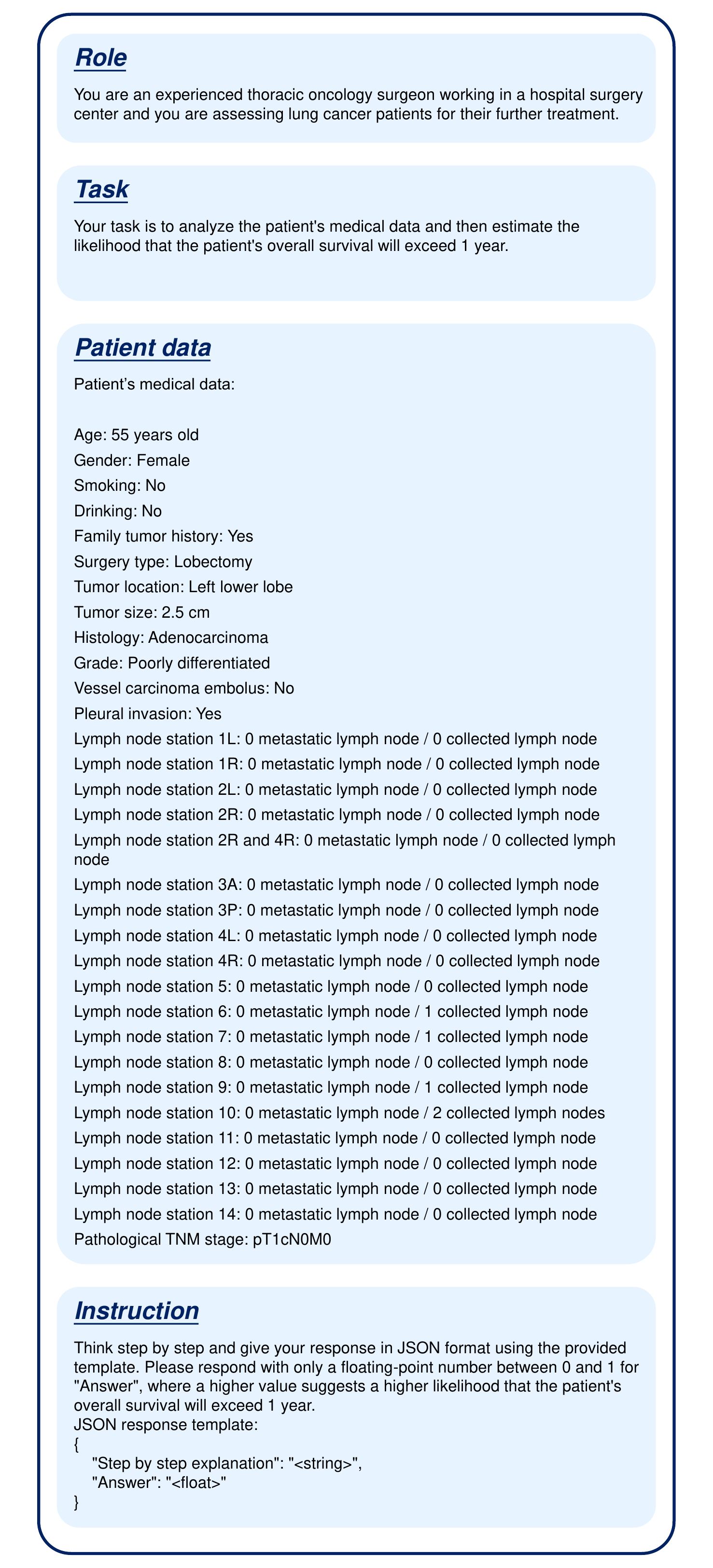}
        \caption{}
    \end{subfigure}
    \begin{subfigure}{0.45\textwidth}
        \includegraphics[width=\textwidth]{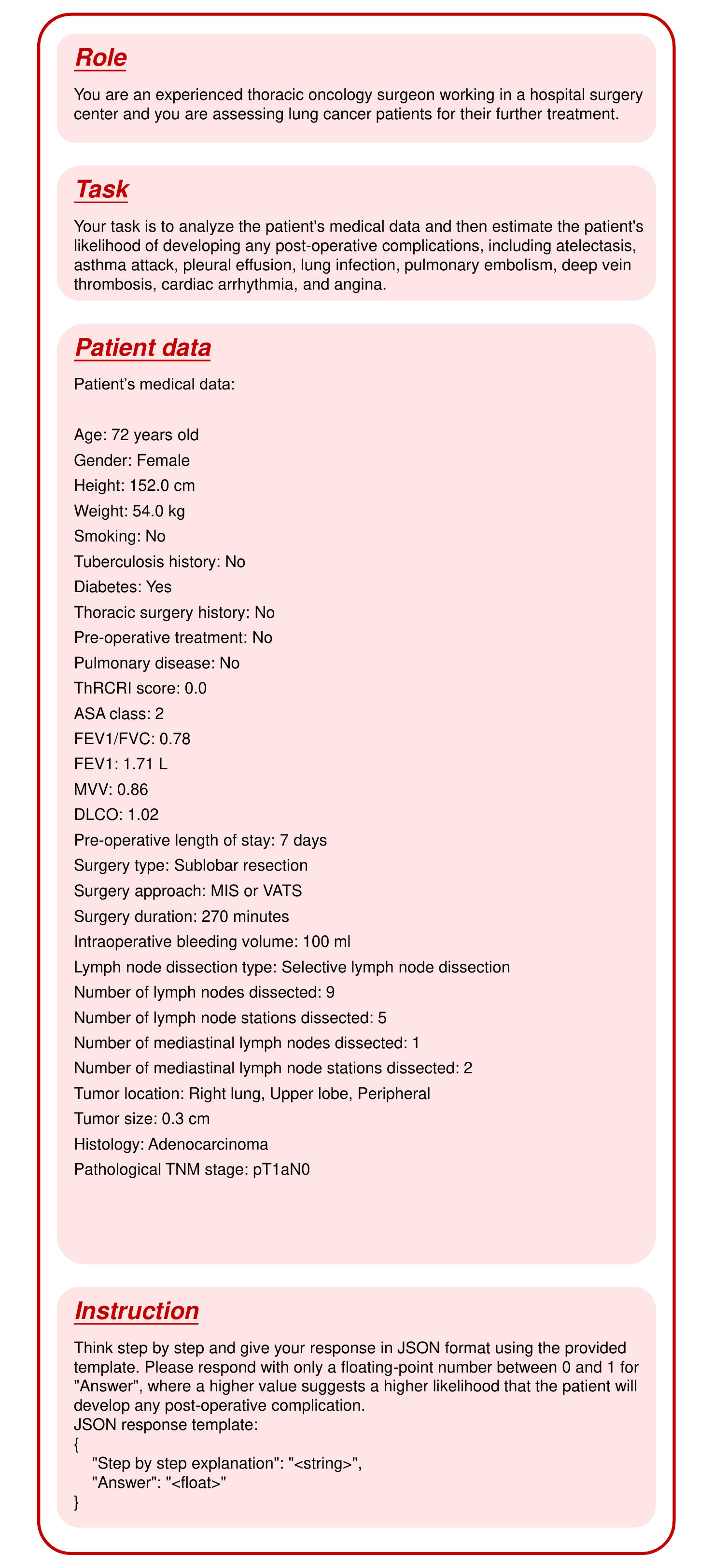}
        \caption{}
    \end{subfigure}
       
    \caption{Prompt templates. (a) 1-year survival prediction prompt template, (b) Combined complication prediction prompt template}
    \label{figure2}
\end{centering}
\end{figure}

\subsection{Prompt design}

The prompt templates of survival and post-operative complication prediction are shown in Figure \ref{figure2}. They consist of 4 elements, i.e., Role, Task, Patient data, and Instruction.

\begin{itemize}
\item \textbf{Role} This element defines the role that LLMs should adopt to generate responses for specific tasks. In this study, we instructed the LLMs to act as thoracic surgeons, responsible for assessing a patient's prognosis and determining the appropriate treatment plan.

\item \textbf{Task} This element specifies the prognosis prediction tasks assigned to the LLMs. For survival prediction, the LLMs were instructed to estimate the likelihood that a patient's survival would exceed $N$ year(s) where $N = \{1,2,3,4,5\}$. For complication prediction, the LLMs were tasked with predicting the likelihood of a patient developing one specific complication or any of the eight complications in our dataset.

\item \textbf{Patient data} This element details the patient clinical data used for evaluation by the large language models (LLMs). For survival prediction, we provided information on patient demographics, medical history, surgical information, pathological characteristics of the primary tumor and lymph nodes, and the pathological TNM stage. For complication prediction, the data included patient demographics, disease history, lung function tests, surgical information, histology, and the pathological TNM stage. The specific data items are illustrated in Figures 2(a) and 2(b).

\item \textbf{Instruction} In this element, we instructed the LLMs to use the Chain-of-Thought strategy to reason step by step in estimating likelihoods. Additionally, the LLMs were directed to provide their responses in JSON format with key-value pairs, such as {"Step by step explanation":"<string>", "Answer":"<float>"}.

\end{itemize}

\subsection{Experimental setup}

In this study, we employed GPT-4o mini (version gpt-4o-mini-2024-07-18) and GPT-3.5 (version gpt-3.5-turbo-1106) to predict the prognoses of lung cancer patients, accessed via the OpenAI API.

For survival prediction, we selected patients who had sufficient follow-up duration for each prediction task. For instance, in the 1-year survival prediction, we included patients who either experienced the event or had no event but were censored after more than one year. Based on the timing of events and censoring, we then assigned 1-year survival labels to the selected patients.

To provide a baseline comparison, we also developed logistic regression (LR) models using the collected datasets. For survival prediction, we implemented a 10-fold cross-validation strategy to split the survival dataset into training-validation and test sets. Within each fold iteration, we performed an additional 5-fold cross-validation on the 9-fold training-validation set to optimize hyperparameters, followed by retraining the model on this set using the selected hyperparameters. The trained model was then evaluated on the 1-fold test set. For complication prediction, we applied the same training, validation, and testing approach as used for survival prediction when predicting combined complications. However, when predicting specific complications, we used a 5-fold cross-validation to split the dataset, followed by a 3-fold cross-validation for hyperparameter tuning, due to the limited number of positive samples for some complications.

We evaluated model performance using two metrics: the area under the receiver operating characteristic curve (AUROC) and the area under the precision-recall curve (AUPRC).

\section{Results}

\subsection{Prognosis data}

In short-term survival prediction tasks, the number of patients who experienced events is relatively small, resulting in data imbalance. For long-term survival prediction tasks, fewer patients had sufficiently long follow-up periods, leading to a smaller sample size. As more events occur over longer periods, the issue of data imbalance is alleviated. Table \ref{table1} presents the statistics of the survival data.

Table \ref{table2} displays the statistics for the post-operative complication data. Complications such as asthma attacks, pulmonary embolism, and deep vein thrombosis were observed in only a few patients, resulting in a significant data imbalance.

\begin{table}[t]
    \caption{The statistics of the survival data.}
    \scriptsize
    \setlength{\tabcolsep}{4pt}
    \centering
    \begin{tabular}{llllll}
    \toprule
         Clinical feature                     & 1-year survival (1277) & 2-year survival (1122) & 3-year survival (970) & 4-year survival (815) & 5-year survival (669)\\
    \midrule
         Age (Mean)                           & 60.14                  & 60.12                  & 60.20                 & 60.24                 & 60.41     \\
         Gender (Male/Female)                 & 794 / 483              & 701 / 421              & 608 / 362             & 512 / 303             & 424 / 245   \\
         Smoking history (Yes/No)             & 698 / 588              & 610 / 512              & 534 / 436             & 445 / 370             & 368 / 301  \\
         Drinking history (Yes/No)            & 364 / 913              & 311 / 811              & 272 / 698             & 230 / 585             & 194 / 475    \\
         Family history (Yes/No)              & 179 / 1098             & 153 / 969              & 134 / 836             & 105 / 710             & 87 / 582  \\
         Surgery type                         &                        &                        &                       &                       &                  \\
         \ \ \ \ Wedge resection              & 40                     & 38                     & 35                    & 30                    & 22     \\
         \ \ \ \ Segmentectomy                & 2                      & 2                      & 1                     & 1                     & 0  \\
         \ \ \ \ Lobectomy                    & 1109                   & 964                    & 833                   & 707                   & 579  \\
         \ \ \ \ Bilobectomy                  & 83                     & 80                     & 68                    & 50                    & 44     \\
         \ \ \ \ Pneumonectomy                & 39                     & 34                     & 31                    & 25                    & 22  \\
         \ \ \ \ Others                       & 4                      & 4                      & 2                     & 2                     & 2     \\
         Tumor location                       &                        &                        &                       &                       &          \\
         \ \ \ \ RUL                          & 415                    & 372                    & 313                   & 261                   & 217     \\
         \ \ \ \ RML                          & 79                     & 68                     & 61                    & 45                    & 41     \\
         \ \ \ \ RLL                          & 258                    & 229                    & 205                   & 174                   & 146     \\
         \ \ \ \ LUL                          & 283                    & 246                    & 209                   & 179                   & 140     \\
         \ \ \ \ LLL                          & 204                    & 178                    & 157                   & 134                   & 108     \\
         \ \ \ \ RC                           & 22                     & 17                     & 16                    & 15                    & 13     \\
         \ \ \ \ LC                           & 16                     & 12                     & 9                     & 7                     & 4     \\
         Tumor size (Mean)                    & 3.17                   & 3.20                   & 3.23                  & 3.28                  & 3.40  \\
         Histology                            &                        &                        &                       &                       &      \\
         \ \ \ \ Adenocarcinoma               & 805                    & 702                    & 611                   & 514                   & 419   \\
         \ \ \ \ Squamous cell carcinoma      & 403                    & 353                    & 298                   & 251                   & 210   \\
         \ \ \ \ Adenosquamous carcinoma      & 17                     & 17                     & 15                    & 12                    & 11   \\
         \ \ \ \ Large cell carcinoma         & 24                     & 24                     & 24                    & 22                    & 17   \\
         \ \ \ \ Others                       & 28                     & 26                     & 22                    & 16                    & 12   \\
         Grade                                &                        &                        &                       &                       &    \\
         \ \ \ \ Poorly differentiated        & 506                    & 455                    & 408                   & 353                   & 307  \\
         \ \ \ \ Moderately differentiated    & 549                    & 503                    & 426                   & 346                   & 275  \\
         \ \ \ \ Well differentiated          & 222                    & 164                    & 136                   & 116                   & 87  \\
         Vessel carcinoma embolus (Yes/No)    & 185 / 1092             & 165 / 957              & 136 / 834             & 124 / 691             & 104 / 565 \\
         Pleural invasion (Yes/No)            & 508 / 769              & 469 / 653              & 430 / 540             & 355 / 460             & 293 / 376   \\
         NLN station 1L (Metastatic/Collected) & 0 / 9                  & 0 / 7                  & 0 / 7                 & 0 / 4                 & 0 / 0  \\
         NLN station 1R (Metastatic/Collected) & 18 / 233               & 18 / 228               & 18 / 212              & 18 / 209              & 18 / 176   \\
         NLN station 2L (Metastatic/Collected) & 1 / 4                  & 1 / 3                  & 1 / 2                 & 1 / 2                 & 1 / 2  \\
         NLN station 2R (Metastatic/Collected) & 67/1057                & 62 / 967               & 60 / 821              & 51 / 727              & 50 / 657  \\ 
         NLN station 2R+4R (Metastatic/Collected)& 151 / 1811           & 128 / 1534             & 114 / 1210            & 104 / 907             & 96 / 638  \\
         NLN station 3A (Metastatic/Collected) & 35 / 488               & 33 / 462               & 32 / 431              & 32 / 359              & 31 / 281   \\
         NLN station 3P (Metastatic/Collected) & 12 / 160               & 12 / 154               & 8 / 127               & 8 / 109               & 8 / 84     \\
         NLN station 4L (Metastatic/Collected) & 38 / 541               & 30 / 483               & 27 / 442              & 22 / 402              & 19 / 322  \\
         NLN station 4R (Metastatic/Collected) & 162 / 1723             & 155 / 1633             & 150 / 1489            & 141 / 1372            & 135 / 1235   \\
         NLN station 5 (Metastatic/Collected)  & 64 / 903               & 55 / 791               & 50 / 663              & 48 / 595              & 45 / 491  \\
         NLN station 6 (Metastatic/Collected)  & 92 /1218               & 86 / 1091              & 67 / 949              & 60 / 820              & 52 / 624  \\
         NLN station 7 (Metastatic/Collected)  & 317 / 5526             & 282 / 4920             & 254 / 4161            & 246 / 3658            & 223 / 2995  \\
         NLN station 8 (Metastatic/Collected)  & 13 /376                & 12 / 359               & 11 / 298              & 11 / 253              & 11 / 233  \\
         NLN station 9 (Metastatic/Collected)  & 27 / 1495              & 27 / 1354              & 25 / 1203             & 25 / 1041             & 23 / 854  \\
         NLN station 10 (Metastatic/Collected) & 222 / 3181             & 208 / 2867             & 188 / 2534            & 177 / 2211            & 167 / 1890  \\
         NLN station 11 (Metastatic/Collected) & 100 / 1630             & 87 / 1406              & 77 / 1233             & 72 / 1047             & 68 / 883  \\
         NLN station 12 (Metastatic/Collected) & 450 / 3501             & 411 / 3162             & 372 / 2757            & 336 / 2335            & 303 / 1986  \\
         NLN station 13 (Metastatic/Collected) & 225 / 1989             & 189 / 1594             & 162 / 1298            & 141 / 1059            & 122 / 790  \\
         NLN station 14 (Metastatic/Collected) & 75 / 768               & 69 / 703               & 62 / 630              & 58 / 590              & 50 / 483  \\
         pT1 (1a/1b/1c)                       & 80 / 316 / 400         & 69 / 277 / 354         & 57 / 235 / 317        & 43 / 199 / 250        & 33 / 147 / 202 /   \\
         pT2 (2a/2b)                          & 228 / 123              & 195 / 112              & 165 / 95              & 149 / 88              & 129 / 80  \\
         pT3                                  & 91                     & 80                     & 70                    & 61                    & 55  \\
         pT4                                  & 39                     & 35                     & 31                    & 25                    & 23  \\
         pN (0/1/2/3)                         & 795 / 224 / 246 / 12   & 693 / 199 / 218 / 12   & 589 / 172 / 197 / 12  & 473 / 149 / 183 / 10  & 372 / 124 / 163 / 10  \\
         Survival status (Alive/Dead)         & 1226 / 51              & 991 / 131              & 770 / 200             & 553 / 262             & 383 / 286 \\
    \bottomrule
    \multicolumn{6}{p{460pt}}{RUL: right upper lobe, RML: right middle lobe, RLL: right lower lobe, LUL: left upper lobe, LLL: left lower lobe, RC: right central, LC: left central, NLN: number of lymph nodes.}\\
    \end{tabular}
    
    \label{table1}
\end{table}

\begin{table}[t]
\scriptsize
\centering
\caption{The statistics of the post-operative complication data}
\setlength{\tabcolsep}{10pt}
\begin{tabular}{llll}
\toprule
Characteristic                             & Statistic        & Characteristic                             & Statistic\\
\midrule
Age (Mean)                                 & 73.53            & NLN dissected (Mean)                       & 14.58   \\
Gender (Male/Female)                       & 323 / 270        & NLNS dissected (Mean)                      & 5.80 \\        
Height (Mean)                              & 162.23           & NMLN dissected (Mean)                      & 7.92 \\
Weight (Mean)                              & 64.75            & NMLNS dissected (Mean)                     & 3.01 \\  
Smoking (Yes/No)                           & 265 / 328        & Tumor location                             &   \\
Tuberculosis history (Yes/No)              & 54 / 539         & \ \ \ \ Up/Middle or Bottom                & 337 / 256  \\  
Diabetes (Yes/No)                          & 107 / 486        & \ \ \ \ Left/Right                         & 263 / 330          \\ 
Thoracic surgery history (Yes/No)          & 38 / 555         & \ \ \ \ Peripheral/Central                 & 539 / 54    \\  
Pre-operative treatment (Yes/No)           & 43 / 550         & Tumor size (Mean)                          & 2.79  \\ 
Pulmonary disease (Yes/No)                 & 35 / 558         & Histology                                  &  \\  
ThRCRI score                               &                  & \ \ \ \ Adenocarcinoma                     & 451   \\  
\ \ \ \ 0                                  & 464              & \ \ \ \ Squamous cell carcinoma            & 115              \\
\ \ \ \ 1-1.5                              & 121              & \ \ \ \ Adenosquamous carcinoma            & 3     \\
\ \ \ \ 2-2.5                              & 0                & \ \ \ \ Large cell carcinoma               & 6          \\
\ \ \ \ >2.5                               & 8                & \ \ \ \ Small cell carcinoma               & 10           \\
ASA class (I/II/III)                       & 28 / 458 / 107   & \ \ \ \ Others                             & 9           \\  
FEV1/FVC (Mean)                            & 0.733            & pT1 (1a/1b/1c)                             & 39 / 136 / 93  \\  
FEV1 (Mean)                                & 3.06             & pT2 (2a/2b)                                & 221 / 30          \\  
MVV (Mean)                                 & 0.80             & pT3                                        & 60           \\  
DLCO (Mean)                                & 0.88             & pT4                                        & 14             \\  
Pre-operative length of stay (Mean)        & 3.9              & pN (x/0/1/2)                               & 44 / 447 / 41 / 61  \\
Surgery type                               &                  & Post-operative complication                &           \\  
\ \ \ \ Wedge                              & 84               & \ \ \ \ Atelectasis                        & 44   \\
\ \ \ \ Sublobar                           & 22               & \ \ \ \ Asthma attack                      & 9        \\
\ \ \ \ Lobectomy                          & 484              & \ \ \ \ Pleural effusion                   & 18     \\
\ \ \ \ Others                             & 3                & \ \ \ \ Lung infection                     & 42   \\
Surgery approach                           &                  & \ \ \ \ Pulmonary embolism                 & 6              \\
\ \ \ \ MIS or VATS                        & 296              & \ \ \ \ Deep vein thrombosis               & 6                 \\
\ \ \ \ Thoracotomy                        & 297              & \ \ \ \ Arrhythmia                         & 88                  \\
Operation duration (Mean)                  & 155.99           & \ \ \ \ Angian                             & 12              \\
Intraoperative bleeding volume (Mean)      & 100.11           & \ \ \ \ Combined                           & 178                \\
Lymph node dissection type                 &                  &\\
\ \ \ \ Selective                          & 414              &\\
\ \ \ \ Systematic                         & 135   \\
\ \ \ \ None                               & 44    \\
\bottomrule

\multicolumn{4}{p{340pt}}{ThRCRI: thoracic revised cardiac risk index, ASA: American Society of Anesthesiologists, FEV1: forced expiratory volume in the first second, FVC: forced vital capacity, MVV: maximal voluntary ventilation, DLCO: diffusing capacity of the lungs for carbon monoxide, MIS: minimally invasive surgery, VATS: video-assisted thoracoscopic surgery, NLN: number of lymph nodes, NLNS: number of lymph node stations,  NMLN: number of mediastinal lymph nodes, NMLNS: number of mediastinal lymph node stations.}\\

\end{tabular}

\label{table2}
\end{table}

\subsection{Predictive performance}

\begin{figure}[h]
\begin{centering}
    \begin{subfigure}{0.48\textwidth}
        \includegraphics[width=\textwidth]{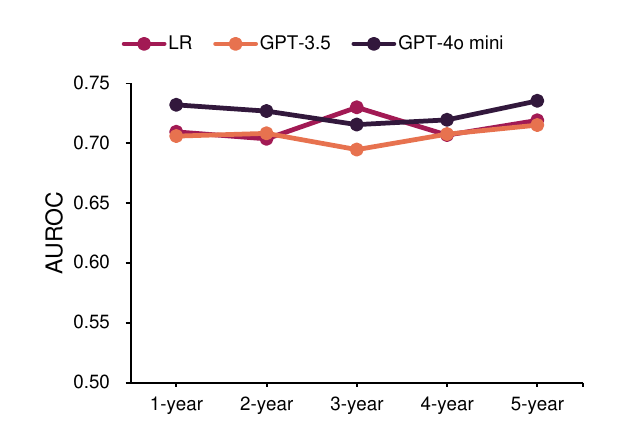}
        \caption{}
    \end{subfigure}
    \begin{subfigure}{0.48\textwidth}
        \includegraphics[width=\textwidth]{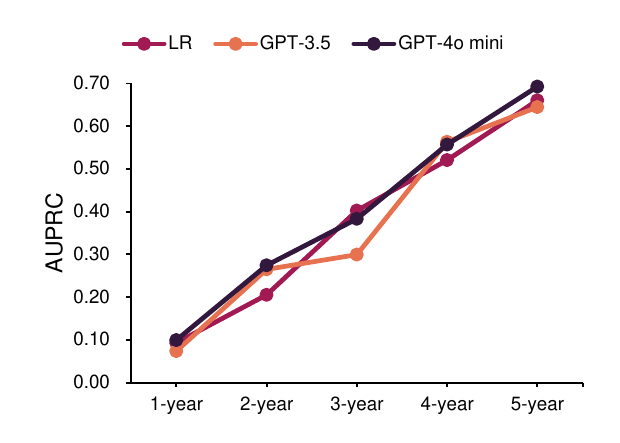}
        \caption{}
    \end{subfigure}
       
    \caption{The AUROC (a) and AUPRC (b) values of the survival prediction models.}
    \label{figure3}
\end{centering}
\end{figure}

Table \ref{table3} presents the predictive performance of the LR, GPT-3.5, and GPT-4o mini models for the survival prediction tasks. GPT-4o mini consistently outperformed GPT-3.5 and the LR models in terms of AUROC values, except for the 3-year survival prediction, where the LR model showed slightly better results. Regarding AUPRC, GPT-4o mini also delivered competitive performance, particularly in the 1-year, 2-year, and 5-year survival predictions, compared to LR and GPT-3.5. On average, GPT-4o mini achieved the highest predictive performance, improving the AUROC by 1.4\% and the AUPRC by 2.5\%. Figure \ref{figure3} provides a visual comparison of the models' AUROC and AUPRC values.

\begin{figure}[h]
\begin{centering}
    \begin{subfigure}{0.48\textwidth}
        \includegraphics[width=\textwidth]{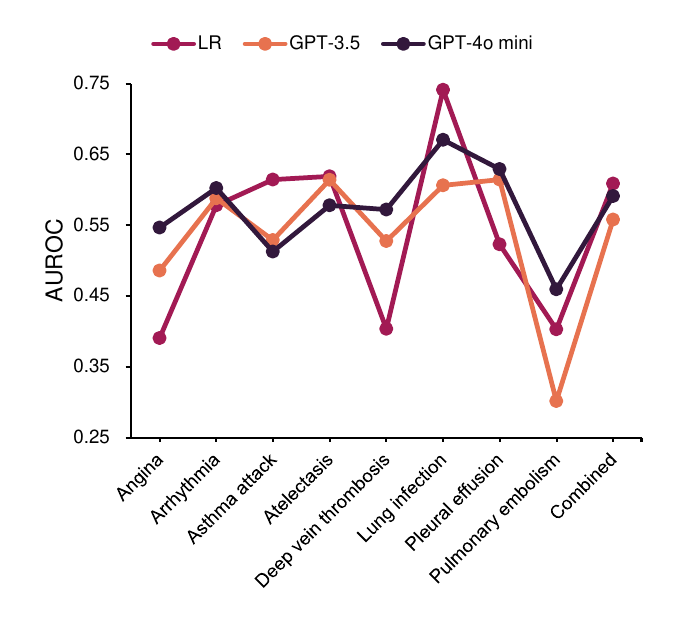}
        \caption{}
    \end{subfigure}
    \begin{subfigure}{0.48\textwidth}
        \includegraphics[width=\textwidth]{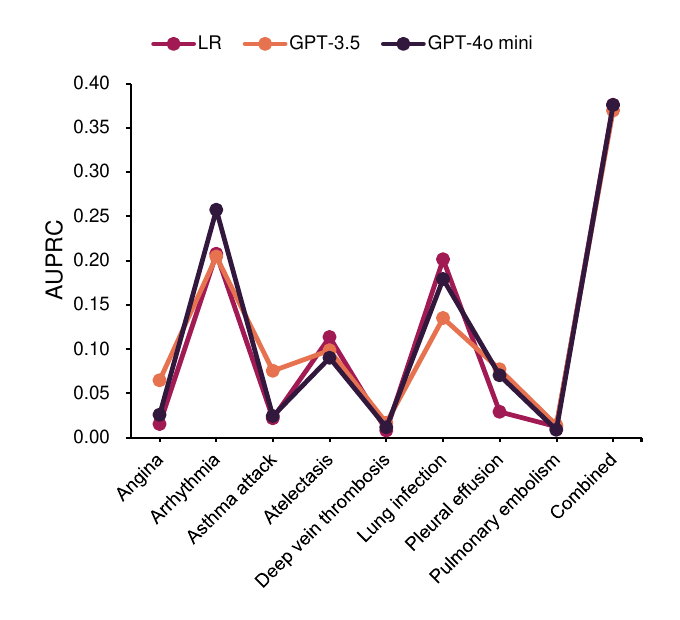}
        \caption{}
    \end{subfigure}
       
    \caption{The AUROC (a) and AUPRC (b) values of the post-operative complication prediction models.}
    \label{figure4}
\end{centering}
\end{figure}

For the post-operative complication prediction tasks, the GPT-4o mini model achieved the highest AUROC values in 5 tasks, while the LR model performed best in the remaining 4 tasks. In terms of AUPRC, GPT-3.5 delivered the best results in 5 prediction tasks, with the LR model leading in 3 tasks. Although GPT-4o mini achieved the highest AUPRC only in the asthma attack prediction, its average AUPRC value was 0.116, just 0.1\% lower than GPT-3.5's best average of 0.117. In terms of AUROC, GPT-4o mini improved results by 3.1\% compared to the LR and GPT-3.5 models. Note that the LR model showed significant performance drops in predicting atelectasis, pulmonary embolism, and angina, with AUROC values falling below 0.5. In contrast, GPT-4o mini demonstrated a more robust performance, with only the AUROC for angina dropping below 0.5. Table \ref{table4} presents the predictive performance of the LR, GPT-3.5, and GPT-4o mini models for the post-operative complication prediction tasks. Figure \ref{figure4} illustrates the results more intuitively.

\begin{table}
    \centering
    \caption{The AUROC and AUPRC values of the LR, GPT-3.5 and GPT-4o mini models for survival prediction.}
    \setlength{\tabcolsep}{10pt}
    \begin{tabular}{lllllll}
    \toprule
    \multirow{2}{*}{Survival prediction tasks} & \multicolumn{3}{l}{AUROC}       & \multicolumn{3}{l}{AUPRC}\\
                                               & LR       & GPT-3.5 & GPT-4o mini                 & LR       & GPT-3.5 & GPT-4o mini           \\ 
    \midrule
    1-year survival prediction                 & 0.710	& 0.706	& \textbf{0.732}	& 0.093	& 0.073	& \textbf{0.099}\\
    2-year survival prediction                 & 0.704	& 0.708	& \textbf{0.727}	& 0.205	& 0.265	& \textbf{0.274}\\
    3-year survival prediction                 & \textbf{0.730}	& 0.695	& 0.716	& \textbf{0.402}	& 0.299	& 0.383\\
    4-year survival prediction                 & 0.707	& 0.708	& \textbf{0.720}	& 0.520	& \textbf{0.563}	& 0.557\\
    5-year survival prediction                 & 0.719	& 0.715	& \textbf{0.735}	& 0.660	& 0.645	& \textbf{0.693}\\
    Average                                    & 0.714	& 0.706	& \textbf{0.726}	& 0.376	& 0.369	& \textbf{0.401}\\
    \bottomrule
    \end{tabular}
    \label{table3}
\end{table}

\begin{table}
    \centering
    \caption{The AUROC and AUPRC values of the LR, GPT-3.5 and GPT-4o mini models for post-operative complication prediction.}
    \setlength{\tabcolsep}{10pt}
    \begin{tabular}{lllllll}
    \toprule
    \multirow{2}{*}{Survival prediction tasks} & \multicolumn{3}{l}{AUROC}       & \multicolumn{3}{l}{AUPRC}\\
                                               & LR       & GPT-3.5 & GPT-4o mini                 & LR       & GPT-3.5 & GPT-4o mini           \\ 
    \midrule
    Atelectasis                                & 0.390	& 0.486	& \textbf{0.546}	& 0.015	& \textbf{0.065}	& 0.025\\
    Asthma attack                              & 0.578	& 0.588	& \textbf{0.602}	& 0.207	& 0.204	& \textbf{0.256}\\
    Pleural effusion                           & \textbf{0.614}	& 0.529	& 0.513	& 0.022	& \textbf{0.075}	& 0.024\\
    Lung infection                             & \textbf{0.619}	& 0.614	& 0.578	& \textbf{0.113}	& 0.098	& 0.090\\
    Pulmonary embolism                         & 0.403	& 0.527	& \textbf{0.572}	& 0.008	& \textbf{0.017}	& 0.011\\
    Deep vein thrombosis                       & \textbf{0.741}	& 0.606	& 0.670	& \textbf{0.201}	& 0.135	& 0.179\\
    Arrhythmia                                 & 0.523	& 0.614	& \textbf{0.629}	& 0.029	& \textbf{0.077}	& 0.070\\
    Angina                                     & 0.403	& 0.301	& \textbf{0.459}	& 0.012	& \textbf{0.014}	& 0.009\\
    Combined                                   & \textbf{0.608}	& 0.558	& 0.591	& \textbf{0.376}	& 0.369	& 0.375\\
    Average                                    & 0.542	& 0.536	& \textbf{0.573}	& 0.109	& \textbf{0.117}	& 0.116\\
    \bottomrule
    \end{tabular}
    \label{table4}
\end{table}

\section{Discussion}

In this study, we investigated the potential of large language models (LLMs) for predicting the prognosis of lung cancer patients. We focused on two key prediction tasks: survival and post-operative complications. To benchmark the performance of the LLMs, we also developed traditional data-driven models using logistic regression as a baseline for comparison.

Based on our experimental results, it is evident that large language models (LLMs) like GPT-4o mini and GPT-3.5 can achieve competitive, and in some tasks superior, predictive performance for lung cancer prognosis compared to traditional data-driven models. Note that LLMs do not require retrospective patient data to learn the latent patterns between features and outcomes. Instead, they leverage the extensive knowledge acquired from vast corpora to estimate the likelihood of outcomes based solely on the data of the current patient, which is like the decision-making process of a true clinician. No need for retrospective data is a significant advantage of LLMs in predicting clinical outcomes, especially given the value and scarcity of patient data, particularly for outcomes such as survival status and post-operative complications. Given their capability in prognosis prediction, LLMs can be utilized in more scenarios, even when patient data is limited or unavailable. Another critical advantage is that LLMs can provide explanations for their predictions, which is essential for gaining clinicians' trust and facilitating the adoption of AI models in clinical workflows to support decision-making.

While the LLMs outperformed the baseline LR models in lung cancer prognosis prediction, it's essential to acknowledge that their superiority may not generalize to all clinical prediction tasks. The medical knowledge embedded in LLMs may vary across different diseases and clinical problems, which could affect their predictive performance. Moreover, the reasoning capabilities of LLMs differ due to variations in their training corpora and methodologies. In our study, the GPT-4o mini showed slightly better results than GPT-3.5, likely due to its more advanced design, as OpenAI stated. However, careful consideration and further exploration are necessary to determine which LLMs are best suited for particular clinical applications.

Additionally, we were unable to collect image data, such as pathology images, which are particularly valuable for predicting lung cancer prognosis. Although some studies have explored the use of LLMs like GPT-4 for diagnosing diseases from image data, these models have not yet shown strong performance in interpreting real-world medical images \cite{YanZhang2023,Nakao2024,ZhouOng2024,Brin2024}. In future work, we plan to collect relevant image data and investigate how to integrate them with clinical data to enhance the effectiveness of LLMs in clinical prediction tasks.

\section{Conclusion}

In this study, we evaluated the performance of LLMs in predicting the prognosis of lung cancer patients. Leveraging knowledge from extensive corpora, LLMs demonstrated competitive and, in some tasks superior, performance in survival and post-operative complication prediction tasks when compared to traditional data-driven logistic regression models. These results indicate that LLMs can effectively predict lung cancer outcomes without relying on extra retrospective patient data, highlighting their potential to offer a novel paradigm for clinical outcome prediction.

\section*{Acknowledgments}

This work was supported by the Beijing Natural Science Foundation (L222020), the National Key R\&D Program of China (No.2022YFC2406804), the Capital’s funds for health improvement and research (No.2024-1-1023), and the National Ten-thousand Talent Program.

\end{document}